\documentclass[10pt,twocolumn,letterpaper]{article}
\usepackage{cvpr}
\usepackage{times}
\usepackage{epsfig}
\usepackage{graphicx}
\usepackage{amsmath}
\usepackage{amssymb}
\usepackage{xspace}

\usepackage[table]{xcolor}
\usepackage{multirow}
\usepackage{tabularx}
\usepackage{colortbl}
\usepackage{booktabs}

\usepackage{caption}

\usepackage{enumitem}
\setitemize{noitemsep,topsep=0pt,parsep=0pt,partopsep=0pt}

\usepackage[numbers,sort]{natbib}

\usepackage{xspace}
\usepackage{bbold}

\newcommand{\set}[1]{\ensuremath{\mathcal{#1}}}

\newcommand{\indep}{\emph{Ours Independent}\xspace}
\newcommand{\joint}{\emph{Ours Combined}\xspace}
\newcommand{\PQOurs}{PQ$^\dagger$\xspace}

\newcommand{\aka}{\emph{a.k.a.}\xspace }

\newcommand{\siABN}{\textrm{iABN$^\text{sync}$}\xspace}

\definecolor{mapillarygreen}{RGB}{5,203,99}

\renewcommand{\paragraph}[1]{
        \vspace{3pt}
	\noindent\textbf{#1}}

\PassOptionsToPackage{hyphens}{url}\usepackage[pagebackref=true,breaklinks=true,letterpaper=true,colorlinks,bookmarks=false]{hyperref}

\cvprfinalcopy 

\def\cvprPaperID{4003 \vspace{-10pt}} 

\ifcvprfinal\fi
\begin{document}

\captionsetup[table]{position=below}
\title{
\vspace{-25pt}
Seamless Scene Segmentation
\vspace{-10pt}
}

\author{Lorenzo Porzi, Samuel Rota Bul\`o, Aleksander Colovic, Peter Kontschieder\\
Mapillary Research\\
{\tt\small research@mapillary.com}
	\vspace{-10pt}
}

\maketitle

\begin{abstract}
In this work we introduce a novel, CNN-based architecture that can be trained end-to-end to deliver seamless scene segmentation results. Our goal is to predict consistent semantic segmentation and detection results by means of a panoptic output format, going beyond the simple combination of independently trained segmentation and detection models. The proposed architecture takes advantage of a novel segmentation head that seamlessly integrates multi-scale features generated by a Feature Pyramid Network with contextual information conveyed by a light-weight DeepLab-like module. As additional contribution we review the panoptic metric and propose an alternative that overcomes its limitations when evaluating non-instance categories. Our proposed network architecture yields state-of-the-art results on three challenging street-level datasets, \ie Cityscapes, Indian Driving Dataset and Mapillary Vistas.
	\vspace{-10pt}
\end{abstract}

\section{Introduction}
Scene understanding is one of the grand goals for automated perception that requires advanced visual comprehension of tasks like semantic segmentation (\textit{Which semantic category does a pixel belong to?}) and detection or instance-specific semantic segmentation (\textit{Which individual object segmentation mask does a pixel belong to?}). Solving these tasks has large impact on a number of applications, including autonomous driving or augmented reality. Interestingly, and despite sharing some obvious commonalities, both these segmentation tasks have been predominantly handled in a disjoint way ever since the rise of deep learning, while earlier works~\cite{Tu+05,YaoFU12,Tighe2014} already approached them in a joint manner. Instead, independent trainings of models, with separate evaluations using corresponding performance metrics, and final fusion in a post-processing step based on task-specific heuristics have seen a revival.

The work in~\cite{Kirillov18} introduces a so-called \textit{panoptic} evaluation metric for joint assessment of semantic segmentation of \textit{stuff} and instance-specific \textit{thing} object categories, to encourage further research on this topic. Stuff is defined as non-countable, amorphous regions of similar texture or material while things are enumerable, and have a defined shape. Few works have started adopting the panoptic metric in their methodology yet, but reported results remain significantly below the ones obtained from fused, individual models. All winning entries on designated panoptic Segmentation challenges like \eg the Joint COCO and Mapillary Recognition Workshop 2018\footnote{\scriptsize \url{http://cocodataset.org/workshop/coco-mapillary-eccv-2018.html}}, were based on combinations of individual (pre-trained) segmentation and instance segmentation models, rather than introducing streamlined integrations that can be successfully trained from scratch.

The use of separate models for semantic segmentation and detection 
obviously comes with the disadvantage of significant computational overhead. Due to a lack of cross-pollination of models, there is no way of enforcing labeling consistency between individual models. Moreover, we argue that individual models supposedly spend significant amounts of their capacity on modeling redundant information, whereas sensible architectural choices in a joint setting are leading to favorable or on par results, but at much reduced computational costs.

In this work we introduce a novel, deep convolutional neural network based architecture for \textit{seamless scene segmentation}. Our proposed network design aims at jointly addressing the tasks of semantic segmentation and instance segmentation. We present ideas for interleaving information from segmentation and instance-segmentation modules and discuss model modifications over vanilla combinations of standard segmentation and detection building blocks. With our findings, we are able to train high-quality, seamless scene segmentation models without the need of pre-trained recognition models. As result, we obtain a state-of-the-art, single model that jointly produces semantic segmentation and instance segmentation results, at a fraction of the computational cost required when combining independently trained recognition models.

We provide the following contributions in our work:
\begin{itemize}
\item Streamlined architecture based on a single network backbone to generate complete semantic scene segmentation for stuff and thing classes
	\item A novel segmentation head integrating multi-scale features from a Feature Pyramid Network~\cite{Lin2016}, 
	with contextual information provided by a light-weight, DeepLab-inspired module
\item Re-evaluation of the panoptic segmentation metric and refinement for more adequate handling of stuff classes
\item Comparisons of the proposed architecture against individually trained and fused segmentation models, including analyses of model parameters and computational requirements
\item Experimental results on challenging driving scene datasets like Cityscapes~\cite{Cordts2016},  Indian Driving Dataset~\cite{Varma19}, and Mapillary Vistas~\cite{Neuhold2017}, demonstrating 	state-of-the-art performance.
\end{itemize}

\section{Related Works}

\paragraph{Semantic segmentation} is a long-standing problem in computer vision research~\cite{Shotton2007,Brostow2008,Krahenbuhl11,Kontschieder2014a} that has significantly improved over the past five years, thanks in great part to advances in deep learning. The works in~\cite{Badrinarayanan2015,Long2015} have introduced encoder/decoder CNN architectures for providing dense, pixel-wise predictions by taking \eg a fully convolutional approach. The more recent DeepLab~\cite{Chen2016} exploits multi-scale features via parallel filters from convolutions with different dilation factors, together with globally pooled features. Another recent Deeplab extension~\cite{Chen2018ECCV} integrates a decoder module for refining object boundary segmentation results. In~\cite{Chen2018NIPS}, a meta-learning technique for dense prediction tasks is introduced, that learns how to design a decoder for semantic segmentation. The pyramid scene parsing network~\cite{zhao2016pspnet} employs i) a pyramidal pooling module to capture sub-region representations at different scales, followed by upsampling and stacking with respective input features and ii) an auxiliary loss applied after the \textit{conv4} block of a ResNet-101 backbone. The works in~\cite{Yu2016,Yu_2017_CVPR} propose aggregation of multi-scale contextual information using dilated convolutions, which have proven to be particularly effective for dense prediction tasks, and are a generalization of the conventional convolution operator to expand its receptive field. RefineNet~\cite{Lin_2017_CVPR} proposes a multi-path refinement network to exploit multiple abstraction levels of features for enhancing the segmentation quality of high-resolution images. Other works like~\cite{Wu2016,RotNeuKon17cvpr} are addressing the problem of class sample imbalance by introducing loss-guided, pixel-wise gradient reweighting schemes.

\paragraph{Instance-specific semantic segmentation} has recently gained large attention in the field, with early, random-field-based works in~\cite{He2014,Tighe2014}. In~\cite{Hariharan2014} a simultaneous detection and segmentation algorithm is developed that classifies and refines CNN features obtained from regions under R-CNN~\cite{Gir+14} bounding box proposals. The work in~\cite{Hayder+17} emphasizes on refining object boundaries for binary segmentation masks initially generated from bounding box proposals. In~\cite{dai2016instance} a multi-task network cascade is introduced that, beyond sharing features from the encoder in all following tasks, subsequently adds blocks for i) bounding box generation, ii) instance mask generation and iii) mask categorization. Another approach~\cite{Dai2016} introduces instance fully convolutional networks that assemble segmentations from position-sensitive score maps, generated by classifying pixels based on their relative positions. The follow-up work in~\cite{li2016fully} builds upon Faster R-CNN~\cite{Ren+15} for proposal generation and additionally includes position-sensitive outside score maps. InstanceCUT~\cite{Kirillov17} obtains instance segmentations by solving a Multi-Cut problem, taking instance-agnostic semantic segmentation masks and instance-aware, probabilistic boundary masks as inputs, provided by a CNN. The work in~\cite{Arnab17} also introduces an approach where an instance Conditional Random Field (CRF) provides individual instance masks based on exploiting box, global and shape cues as unary potentials, together with instance-agnostic semantic information. In~\cite{Liu17}, sequential grouping networks are presented that run a sequence of simple networks for solving increasingly complex grouping problems, eventually yielding instance segmentation masks. DeepMask~\cite{Pinheiro15} first produces an instance-agnostic segmentation mask for an input patch, which is then assigned to a score corresponding to how likely this patch it to contain an object. At inference, their approach generates a set of ranked segmentation proposals. The follow-up work SharpMask~\cite{Pinheiro16} augments the networks with a top-down refinement approach. Mask R-CNN~\cite{He2017} forms the basis of current state-of-the-art instance segmentation approaches. It is a conceptually simple extension of Faster R-CNN, adding a dedicated branch for object mask segmentation in parallel to the existing ones for bounding box regression and classification. Due to its importance in our work, we provide a more thorough review in the next section. The work in~\cite{Liu2018} proposes to improve localization quality of objects in Mask R-CNN via integration of multi-scale information as bottom-up path augmentation. 

\paragraph{Joint segmentation and instance-segmentation} approaches date back to~\cite{Tu+05}, introducing a Bayesian approach for scene representation by establishing a scene parsing graph to explain both, segmentation of stuff and things. Other works before the era of deep learning often built upon CRFs where \cite{Tighe2014} alternatingly refined pixel labelings and object instance predictions, and~\cite{YaoFU12} framed holistic scene understanding as a structure prediction problem in a graphical model, defined over hierarchies of regions, scene types, \etc. The recently proposed work in~\cite{Kendall18} addresses automated loss balancing in a multi-task learning problem based on analysing the homoscedastic uncertainty of each task.  Even though their work addresses three tasks at the same time (semantic segmentation, instance segmentation and depth estimation), it fails to demonstrate consistent improvements over semantic segmentation and instance segmentation alone and lacks of comparisons to comparable baselines. The supervised variant in~\cite{Li2018Weaklypanoptic} generates panoptic segmentation results but \textit{i)} requires separate (external) input for bounding box proposals and \textit{ii)} exploits a CRF during inference, increasing the complexity of the model. The work in~\cite{Geus18} attempts to introduce a unified architecture related to our ideas, however, the reported results remain significantly below those of reported state-of-the-art methods. Independently and simultaneously to our paper, a number of works~\cite{Li2018,PanopticCOCOWinners2018,Xiong_UBER_2019,Kirillov19,Yang_Google_2019} have proposed panoptic segmentation provided by a single deep network, confirming the importance of this task to the field. While comparable in complexity and architecture, we obtain improved performance on challenging street-level image datasets like Cityscapes and Mapillary Vistas.

\section{Proposed Architecture}
\label{sec:method}

\begin{figure*}
	\centering
	\includegraphics[width=.7\textwidth]{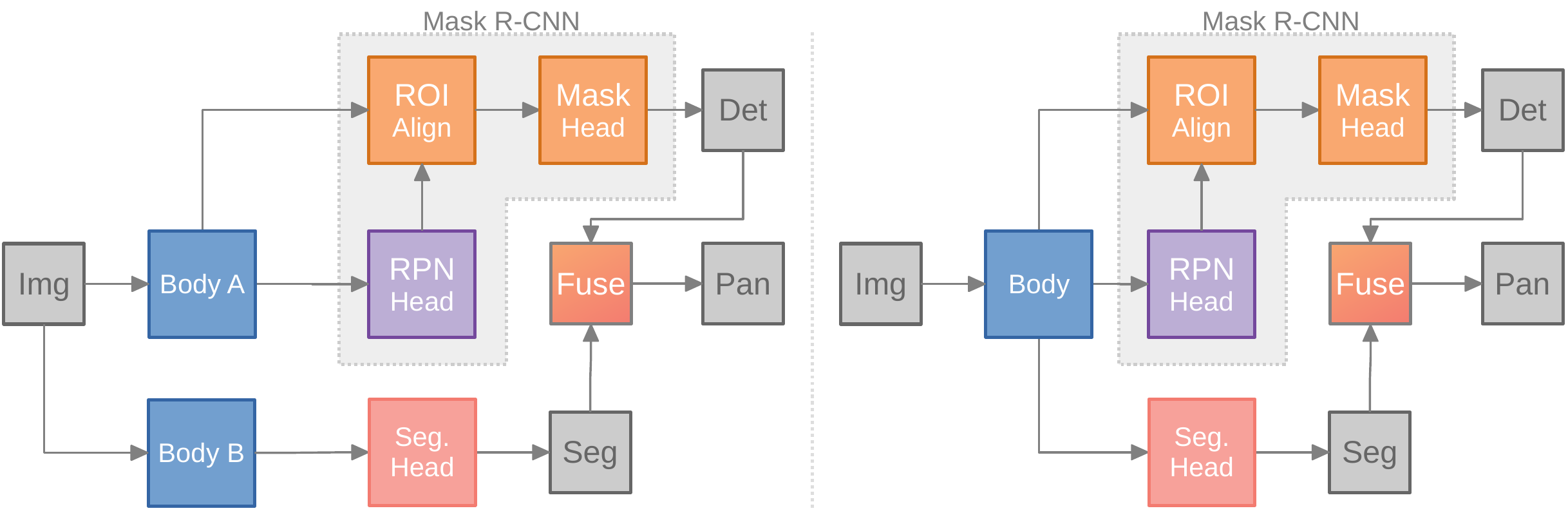}
	\caption{Comparison of two architectures for panoptic segmentation. 
		Left: Separate models (including bodies) for detection and segmentation. Both predictions are fused to obtain the final panoptic prediction.
		Right: Shared body between the heads. 
	} \label{fig:basic}
		\vspace{-10pt}
\end{figure*}

The proposed architecture consists of a backbone working as feature extractor and two task-specific branches addressing semantic segmentation and instance segmentation, respectively. Hereafter, we provide details about each component and refer to Fig.~\ref{fig:basic} for an overview.

\subsection{Shared Backbone}
The backbone that we use throughout this paper is a slightly modified ResNet-50~\cite{He+16} with a Feature Pyramid Network (FPN)~\cite{Lin2016} on top. 
The FPN network is linked to the output of the modules conv2, conv3, conv4 and conv5 of ResNet-50, which yield different downsampling factors, namely $\times 4$, $\times 8$, $\times 16$ and $\times 32$, respectively.
Akin to the original FPN architecture, we have a variable number of additional, lower resolution scales covering downsampling factors of $\times 64$ and $\times 128$, depending on the dataset.
The main modification in ResNet-50 is the replacement of all Batch Normalization (BN) + ReLU layers with synchronized Inplace Activated Batch Normalization (\siABN) proposed in~\cite{RotPorKon18a}, which uses LeakyReLU with slope $0.01$ as activation function due to the need of invertible activation functions. 
This modification gives two important advantages: i) we gain up to 50\% additional GPU memory since the layer performs in-place operations, and ii) the synchronization across GPUs ensures a better estimate of the gradients in multi-GPU trainings with positive effects on convergence.

\subsection{Instance Segmentation Branch}
\label{sec:msk}
The instance segmentation branch follows the state-of-the-art Mask R-CNN~\cite{He2017} architecture.
This branch is structured into a region proposal head and a region segmentation head.

\paragraph{Region Proposal Head (RPH).}
The RPH introduces the notion of an \emph{anchor}. An anchor is a reference bounding box (\aka~region), centered on one of the available spatial locations of the RPH's input and having pre-defined dimensions. The set of pre-defined dimensions is chosen in advance, depending on the dataset and the scale of the FPN output (see details in Sec.~\ref{sec:results}).
We denote by $\set A$ all anchors that can be constructed by combining a position on an available, spatial location and a dimension from the pre-defined set, and which are entirely contained in the image.
Given an anchor $a$ we denote its position (in the image coordinate system) by $(u_a,v_a)$ and its dimensions by $(w_a,h_a)$. 
The role of RPH is to apply a transformation to each anchor in order to obtain a new bounding box proposal together with an objectness score, that assesses the validity of the region.
To this end, RPH applies a $3\times 3$ convolution with $256$ output channels and stride $2$ to the outputs of the backbone, followed by \siABN, and a $1\times 1$ convolution with $5 N_{\text{anchors}}$ channels, which provide a bounding box proposal with an objectness score for each anchor in $\set A$. In more details, for each anchor $a\in\set A$ the transformed bounding box has center $(\hat u,\hat v)=(u_a+o_u w_a,v_a+o_v h_a)$, dimensions $(\hat w,\hat h)=(w_a\,e^{o_w},h_a\,e^{o_h})$ and objectness score $\hat s=\sigma(o_s)$, where $(o_u,o_v,o_w,o_h,o_s)$ represents the output from the $1\times 1$ convolution for anchor $a$, and $\sigma(\cdot)$ is the sigmoid function. 
The resulting set of bounding boxes are then fed to the region segmentation head, with distinct filtering steps for training and test time.

\paragraph{Region Segmentation Head (RSH).}
Each region proposal $\hat r=(\hat u,\hat v, \hat w,\hat h)$ obtained from RPH is fed to RSH, which applies ROIAlign~\cite{He2017}, pooling features directly from the $k$th output of the backbone within region $\hat r$ with a $14\times 14$ spatial resolution, where $k$ is selected based on the scale of $\hat r$ according to the formula $k=\max(1,\min(4,\lfloor 3+\log_2(\sqrt{\hat w\hat h}/224) \rfloor))$~\cite{He2017}.
The result is forwarded to two parallel sub-branches: one devoted
to predicting a class label (or void) for the region proposal together with class-specific corrections of the proposal's bounding box, and the other devoted to providing class-specific mask segmentations. 
The first sub-branch of RSH is composed of two fully-connected layers with 1024 channels, each followed by Group Normalization (GN)~\cite{Wu_2018_ECCV} and LeakyReLU with slope $0.01$, and a final 
fully-connected layer with $5 N_{\text{classes}}+1$ output units.
The output units encode, for each possible class $c$, class-specific correction factors $(o_u^c,o_v^c,o_w^c,o_h^c)$ that are used to compute a new bounding box centered in $(\hat u^c, \hat v^c)=(\hat u+o_u^c\hat w,\hat v+o_v^c\hat h)$ with dimensions $(\hat w^c, \hat h^c)=(\hat w\,e^{o_w^c},\hat h\,e^{o_h^c})$. This operation generates from $\hat r$ and for each class $c$ a new class-specific region proposals given by $\hat r^c=(\hat u^c, \hat v^c, \hat w^c,\hat h^c)$.
In addition, we have $N_\text{classes}+1$ units providing logits for a softmax layer that gives a probability distribution over classes and void, the latter label assessing the invalidity of the proposal. The probability associated to class $c$ is used as score function $\hat s^c$ for the class-specific region proposal $\hat r^c$.
The second sub-branch applies four $3\times 3$ convolution layers each with $256$ output channels. As for the first sub-branch each convolution is followed by GN and LeakyReLU.
This is followed by a $2\times 2$ deconvolution layer with output stride $2$ and $256$ output channels, GN, LeakyReLU, and a final $1\times 1$ convolution with $N_\text{classes}$ output channels. This yields, for each class, $28\times 28$ logits that provide class-specific mask foreground probabilities for the given region proposal via a sigmoid.
The resulting mask prediction is combined with the output of the segmentation branch described below. 

\subsection{Semantic Segmentation Branch}
\label{sec:sem}
\begin{figure}[ht]
	\centering
	\includegraphics[width=.7\columnwidth]{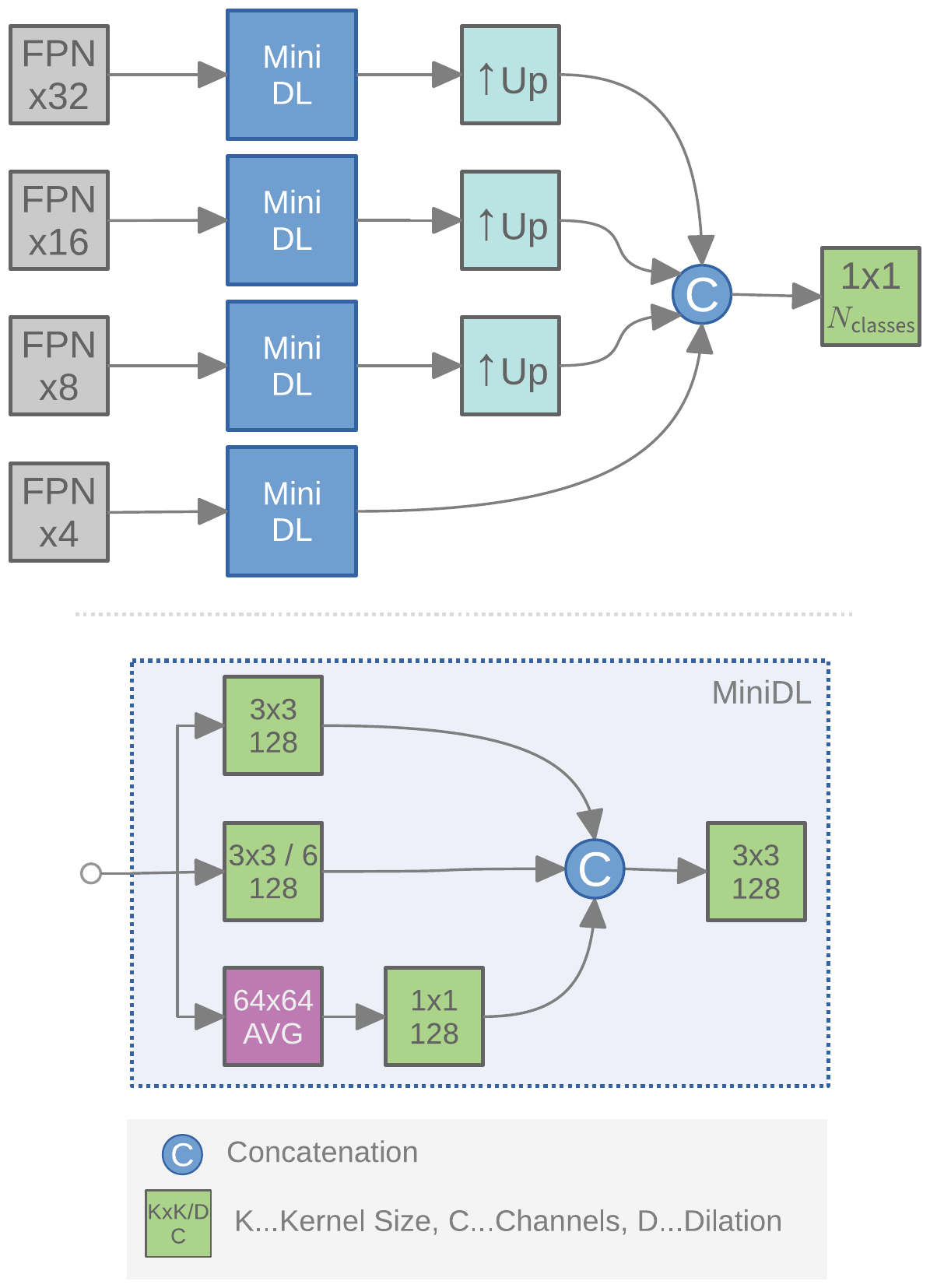}
	\caption{Segmentation Head (top) and the architecture of the Mini Deeplab (MiniDL) module (bottom), which is used in the head.}
	\label{fig:minidl}
	\vspace{-10pt}
\end{figure}

The semantic segmentation branch takes as input the outputs of the backbone corresponding to the first four scales of FPN. We apply independently to each input (not sharing parameters) a variant of the DeepLabV3 head~\cite{Chen2017} that we call \emph{Mini-DeepLab} (MiniDL, see Fig.~\ref{fig:minidl}) followed by an upsampling operation that yields an output downsampling factor of $\times 4$ and $128$ output channels. All the resulting streams are concatenated and the result is fed to a final $1\times 1$ convolution layer with $N_\text{classes}$ output channels. The output 
is bilinearly upsampled to the size of the input image. This provides the logits for a final softmax layer that provides class probabilities for each pixel of the input image.
Each convolution in the semantic segmentation branch, including MiniDL, is followed by \siABN akin to the backbone.

\paragraph{MiniDL.}
The MiniDL module consists of $3$ parallel sub-branches. The first two apply a $3\times 3$ convolution with $128$ output channels with dilations $1$ and $6$, respectively. The third one applies a $64\times 64$ average pooling operation with stride $1$ followed by a padding with boundary replication to recover the spatial resolution of the input and a $1\times 1$ convolution with $128$ output channels. The outputs of the $3$ sub-branches are concatenated and fed into a $3\times 3$ convolution layer with $128$ output channels, which delivers the final output of the MiniDL module. 

\paragraph{} As opposed to DeepLabV3, we do not perform the global pooling operation in our MiniDL module for two reasons: i) it breaks translation equivariance if we change the input resolution at test time, which is typically the case and ii) since we work with large input resolutions, it is preferable to limit the extent of contextual information. Instead, we replaced the global pooling operation with average pooling in the $3$rd sub-branch with a fixed large kernel size and stride $1$, but without padding. The lack of padding yields an output resolution which is smaller than the input resolution and we re-establish the input resolution by replicating the boundary of the resulting tensor, \ie we employ a padding layer with boundary replication. By doing so, we generalize the solution originally implemented in DeepLabV3, for we obtain the same output at training time if we keep the kernel size equal to the \emph{training} input resolution, but we preserve translation equivariance at test time, and can reduce the extent of contextual information by properly fixing the kernel size.

\subsection{Training losses}\label{ss:losses}

The two branches of the architecture are supported with distinct losses, which are detailed below.
We denote by $\set Y=\{1,\dots,N_\text{classes}\}$ the set of class labels, and assume for simplicity input images with fixed resolution $H\times W$.

\paragraph{Semantic segmentation branch.}
Let $Y_{ij}\in \set Y$ be the semantic segmentation ground truth for a given image and pixel position $(i,j)$ and let $P_{ij}(c)$ denote the 
predicted probability for the same pixel to be assigned class $c\in\set Y$.
The per-image segmentation loss that we employ is a weighted per-pixel log-loss that is given by
\[
	L_{ss}(P,Y)=-\sum_{ij}\omega_{ij}\log P_{ij}(Y_{ij})\,.
\]
The weights are computed following the simplest version of~\cite{RotNeuKon17cvpr} with $p=1$ and $\tau=\frac{4}{WH}$. This corresponds to having a pixel-wise hard negative mining, which selects the $25\%$ worst predictions, \ie $\omega_{ij}=\tau$ for all $(i,j)$ within the $25\%$ pixels yielding the lowest probability $P_{ij}(Y_{ij})$, and $\omega_{ij}=0$ otherwise.

\paragraph{Instance segmentation branch.}
The losses for the instance segmentation branch and the training procedure are derived from the ones proposed in Mask R-CNN~\cite{He2017}. 
We start with the losses applied to the output of RPH. Let $\set R$ be the set of ground truth bounding boxes for a given image $I$, let $\hat{\set R}$ be the set of bounding boxes generated from $I$ by RPH and let
$\hat r_a\in\hat{\set R}$ be the proposal originated from anchor $a\in\set A$.
We employ the same strategy adopted in Mask-RCNN to assign ground truth boxes in $\set R$ to predictions in $\hat {\set R}$.
For each ground truth box $r\in\set R$, the closest anchor in $\set A$, \ie the one with largest Intersection over Union (IoU), is selected and the corresponding prediction $\hat r_a\in\hat {\set R}$ is regarded as positively matching $r$. For each anchor $a\in\set A$, we seek the closest ground truth box $r\in\set R$ and regard the match between $r$ and $\hat r_a$, \ie the predicted box in $\hat{\set R}$ corresponding to anchor $a$, as positive match if $\text{IoU}(r,a)>\tau_H$, or negative match if $\text{IoU}(r,a)<\tau_L$, where $\tau_H>\tau_L$ are user-defined thresholds (we use $\tau_H=0.7, \tau_L=0.3$). We take a random subset $\set M_+\subset \set R\times\hat{\set R}$ of all positive matches and a random subset $\set M_-\subset \set R\times\hat{\set R}$ of all negative matches, where we typically set $|\set M_+|\leq 128$ and $|\set M_-|\leq 256-|\set M_+|$ in order to have at most $256$ matches. 
The objectness loss for the given image is
\begin{align*}
	L^\text{ob}_\text{RPH}(\set M_{\pm}) =& -\frac{1}{\left|\set M\right|} \sum_{(r,\hat r)\in\set M_+}\log s_{\hat r} \\
	& -\frac{1}{\left|\set M\right|} \sum_{(r,\hat r)\in\set M_-}\log (1- s_{\hat r})\,,
\end{align*}
where $s_{\hat r}$ is the objectness score of the predicted bounding box $\hat r$, and $\set M = \set M_+ \cup \set M_-$, while
the bounding box regression loss is defined only on positive matches $\set M_+$ and is given by
\begin{multline*}
	L^\text{bb}_\text{RPH}(\set M_\pm)=\frac{1}{\left|\set M\right|} \sum_{(r,\hat r)\in\set M_+}\left\{\left|\frac{x_r-x_{\hat r}}{w_{a_{\hat r}}}\right|_S\right.\\
	\left.+\left|\frac{y_r-y_{\hat r}}{h_{a_{\hat r}}}\right|_S +\left|\log \frac{w_{\hat r}}{w_r}\right|_S+\left|\log \frac{h_{\hat r}}{h_r}\right|_S\right\}\,,
\end{multline*}
where $|\cdot|_S$ is the smooth $L_1$ norm~\cite{Ren+15}, $r=(x_r,y_r,w_r,h_r)$, $\hat r=(x_{\hat r},y_{\hat r},w_{\hat r},h_{\hat r})$ and $a_{\hat r}$ is the anchor that originated prediction $\hat r$.

We next move to losses that pertain to RSH.
Let again $\set R$ be the ground truth boxes for a given image $I$ and let $\breve {\set R}$ be the union of $\set R$ and the set of bounding boxes generated by RPH from $I$, filtered with Non-Maxima Suppression (NMS), and clipped to the image area. For each $\breve r\in\breve{\set R}$, we find the closest (in terms of IoU) ground truth box $r\in\set R$ and regard it as a positive match if $\text{IoU}(r,\breve r)>\eta$ and as negative match otherwise (we use $\eta=0.5$). Let $\set N_+$ and $\set N_-$ be random subsets or all positive and negative matches, respectively, where we typically set $|\set N_+|\leq 128$ and $|\set N_-|\leq 512-|\set N_+|$ in order to have at most $512$ matches.
The region proposal classification loss is given by
\begin{align*}
	L^\text{cls}_\text{RSH}(\set N_\pm) = &-\frac{1}{\left|\set N\right|} \sum_{(r,\breve r)\in\set N_+} \log s_{\breve r}^{c_r} \\
	&- \frac{1}{\left|\set N\right|} \sum_{(r,\breve r)\in\set N_-} \log s_{\breve r}^\emptyset\,,
\end{align*}
where $\set N = \set N_+ \cup \set N_-$, $\emptyset$ denotes the void class, $c_r$ is the class of the ground truth bounding box $r\in\set R$, $s_{\breve r}^c$ is the probability given by RSH of the input proposal $\breve r$ to take class $c$.
The bounding box regression loss is defined only on positive matches $\set N_+$ and is given by
\begin{multline*}
	L^\text{bb}_\text{RSH}(\set N_\pm)=\frac{1}{\left|\set N\right|} \sum_{(r,\breve r)\in\set N_+}\left\{\left|\frac{x_r-x^{c_r}_{\breve r}}{w_{\breve r}}\right|_S\right.\\
	\left.+\left|\frac{y_r-y^{c_r}_{\breve r}}{h_{\breve r}}\right|_S+\left|\log \frac{w^{c_r}_{\breve r}}{w_r}\right|_S+\left|\log \frac{h_{\breve r}^{c_r}}{h_r}\right|_S\right\}\,.
\end{multline*}

Finally, the mask segmentation loss is given as follows.
Let $\breve r$ be a region proposal entering RSH matching a ground-truth region $r$, \ie $(r,\breve r)\in\set N_+$. Let $S^r\in\{0,1,\emptyset\}^{28\times 28}$ be the corresponding ground truth binary mask with associated class label $c$, where $\emptyset$ denotes a void pixel, and let $S^{\breve r}\in[0,1]^{28\times 28}$ be the mask prediction for class $c$ obtained from RSH with entries $S^{\breve r}_{ij}$ denoting the probability of cell $(i,j)$ to belong to the instance conditioned on class $c$ and region $\breve r$. Then, the loss for the matched proposal $\breve r$ is given by
\begin{align*}
	L^\text{msk}_\text{RSH}(S^r,S^{\breve r})&=-\frac{1}{|\set P^r|}\sum_{(i,j)\in\set P^r}S^r_{ij}\log S^{\breve r}_{ij} \\
	&\quad- \frac{1}{|\set P^r|}\sum_{(i,j)\in\set P^r}(1-S^r_{ij})\log(1-S^{\breve r}_{ij})\,,
\end{align*}
where $\set P^r$ is the set of pixels being non-void in the ground-truth $28\times 28$ mask $S^r$.
The mask losses corresponding to all matched proposals are finally summed and divided by $|\set N|$ akin to the bounding box loss $L^\text{bb}_\text{RSH}$.

All losses are weighted equally and gradient from RSH flows only to the backbone, \ie the gradient component flowing from RSH to RPH is blocked, akin to Mask-RCNN.

\subsection{Testing and Panoptic Fusion}\label{ssec:fusion}
 At test time, given an input image $I$ we extract features $F$ with the backbone and generate region proposals with corresponding objectness scores by applying RPH. We filter the resulting set of bounding boxes with Non-Maxima Suppression (NMS)
guided by the objectness scores. The surviving proposals are fed to the RSH (first sub-branch) together with $F$ in order to generate class-specific region proposals with corresponding class probabilities. A second NMS pass is applied on the resulting set of bounding boxes, this time independently per class guided by the class probabilities. The resulting class-specific bounding boxes are fed again to RSH together with $F$, but this time through the second sub-branch which provides the corresponding mask predictions. 
The extracted features $F$ are fed in parallel to the segmentation branch, which provides class probabilities for each pixel. 
The output of RSH and the segmentation branch are finally fused using the strategy given below, in order to deliver the final panoptic segmentation.

\paragraph{Fusion.} 
The fusion operation is inspired by the one proposed in~\cite{Kirillov18}.
We start iterating over predicted instances in reverse classification score order. For each instance we mark the pixels in the final output that belong to it and are still unassigned, provided that the latter number of pixels covers at least $50\%$ of the instance. Otherwise we discard the instance thus resembling a NMS procedure.
Remaining unassigned pixels take the most likely class according to the segmentation head prediction, if it belongs to stuff, or void if it belongs to thing. Finally, if the total amount of pixels of any stuff class is smaller than a given threshold ($4096$ in our case) we mark all those pixels to void.


\section{Revisiting Panoptic Segmentation}
In this section we review the panoptic segmentation metric~\cite{Kirillov18} (\aka~PQ metric), which evaluates the performance of a so-called panoptic segmentation, and discuss a limitation of this metric when it comes to stuff classes.

\paragraph{PQ metric.} A panoptic segmentation assigns each pixel a stuff class label or an instance ID. Instance IDs are further given a thing class label (\eg pedestrian, car, \etc). As opposed to AP metrics used in detection, instances are not overlapping. The PQ metric is computed for each class independently and averaged over classes (void class excluded). This makes the metric insensitive to imbalanced class distributions. Given a set of ground truth segments $\set S_c$ and predicted segments $\hat {\set S}_c$ for a given class $c$, the metric collects a set of True Positive matches as $\text{TP}_c=\{(s,\hat s)\in\set S_c\times\hat{\set S}_c\,:\, \text{IoU}(s,\hat s)>0.5\}$\,.
This set contains all pairs of ground truth and predicted segments that overlap in terms of IoU more than $0.5$. By construction, every ground truth segment can be assigned at most one predicted segment and vice versa. 
The PQ metric for class $c$ is given by
\[
	\text{PQ}_c=\frac{\sum_{(s,\hat s)\in \text{TP}_c}\text{IoU}(s,\hat s)}{|\text{TP}_c|+\frac{1}{2}|\text{FP}_c|+\frac{1}{2}|\text{FN}_c|}\,,
\]
where $\text{FP}_c$ is the set False Positives, \ie unmatched predicted segments for class $c$, and $\text{FN}_c$ is the set False Negatives, \ie unmatched segments from ground truth for class $c$.
The metric allows also specification of void classes, both in ground truth and actual predictions. 
Pixels labeled as void in the ground truth are not counted in IoU computations and predicted segments of any class $c$ that overlap with void more than $50\%$ are not counted in $\text{FP}_c$. Also, ground truth segments for class $c$ that overlap with predicted void pixels more than $50\%$ are not counted in $\text{FN}_c$. 
The final PQ metric is obtained by averaging the class-specific PQ scores:
\[
	\text{PQ}=\frac{1}{N_\text{classes}}\sum_{c\in\set Y}\text{PQ}_c\,.
\]
We further denote by $\text{PQ}_\text{Th}$ and $\text{PQ}_\text{St}$ the average of thing-specific and stuff-specific PQ scores, respectively.

\paragraph{The issue with stuff classes.}
\begin{figure}
    \includegraphics[width=\columnwidth]{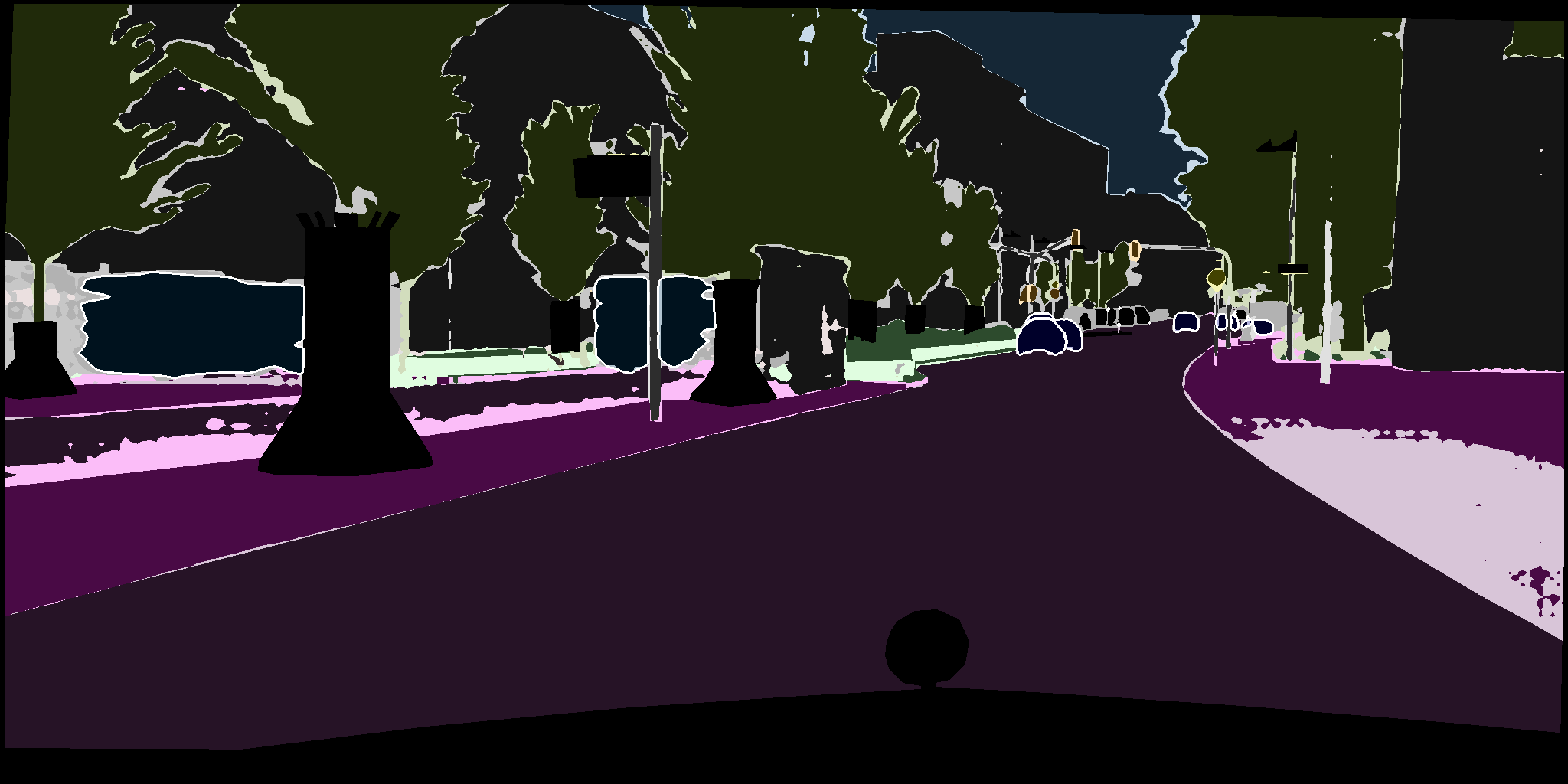}
    \caption{Prediction on a Cityscapes validation set image, where light colored areas highlight conducted errors. Several classes, \eg pole (IoU 0.49) and traffic light (IoU 0.46), are just below the PQ acceptance threshold, while the sidewalk class (IoU 0.62) is just above it. Thus, the former will be overly penalized (PQ $\rightarrow 0$), while the latter will contribute positively (PQ $\rightarrow 0.62$), even if they look qualitatively similar. Best viewed in color and with digital zoom.}
    \label{fig:pq_example}
    \vspace{-10pt}
\end{figure}
One limitation of the PQ metric is that it over-penalizes errors related to stuff classes, which are by definition not organized into instances.
This derives from the fact that the metric does not distinguish stuff and thing classes and applies indiscriminately the rule that we have a true positive if the ground truth and the predicted segment have IoU greater than 0.5. De facto it regards all pixels in an image belonging to a stuff class as a single big instance.
To give an example of why we think this is sub-optimal, consider a street scene with two sidewalks and assume that the algorithm confuses one of the two with road (say the largest) then the segmentation quality on sidewalk for that image becomes $0$. A real-world example is provided in Fig.~\ref{fig:pq_example}, where several stuff segments are severely penalized by the PQ metric, not reflecting the real quality of the segmentation.
The $>$$0.5$-IoU rule for thing classes is convenient because it renders the matching between predicted and ground truth instances easy, but this is a problem to be solved only for thing classes. Indeed, predicted and ground truth segments belonging to stuff classes can be directly matched independently from their IoU because each image has at most one instance of them.

\paragraph{Suggested alternative.}
We propose to maintain the PQ metric only for thing classes, but change the metric for stuff classes.
Specifically, let $\set S_c$ be the set of ground truth segments of a given class $c$ and let $\hat {\set S}_c$ be the set of predicted segments for class $c$.
Note that each image can have at most $1$ ground truth segment and at most $1$ predicted segment of the given stuff class.
Let $\set M_c=\{(s,\hat s)\in \set S_c\times\hat {\set S}_c\,:\, \text{IoU}(s,\hat s)>0\}$ be the set of matching segments, then the updated metric for class $c$ becomes:
\[
	\text{PQ}^\dagger_c = 
	\begin{cases}
		\frac{1}{|\set S_c|}\sum_{(s,\hat s)\in\set M_c}\text{IoU}(s,\hat s)\,,&\text{if $c$ is stuff class}\\
		\text{PQ}_c\,,&\text{otherwise}.
\end{cases}
\]
We denote by $\text{PQ}^\dagger$ the final version of the proposed panoptic metric, which averages $\text{PQ}^\dagger_c$ over all classes, \ie
\[
	\text{PQ}^\dagger=\frac{1}{N_\text{classes}}\sum_{c\in\set Y}\text{PQ}^\dagger_c\,.
\]
Similarly to PQ, the proposed metric is bounded in $[0,1]$ and implicitly regards a stuff segment of an image as a single instance. However, we do not require the prediction of stuff classes to have IoU$>$$0.5$ with the ground truth.

\section{Experimental Results}
\label{sec:results}

\begin{figure*}
	\centering
	\includegraphics[width=\textwidth]{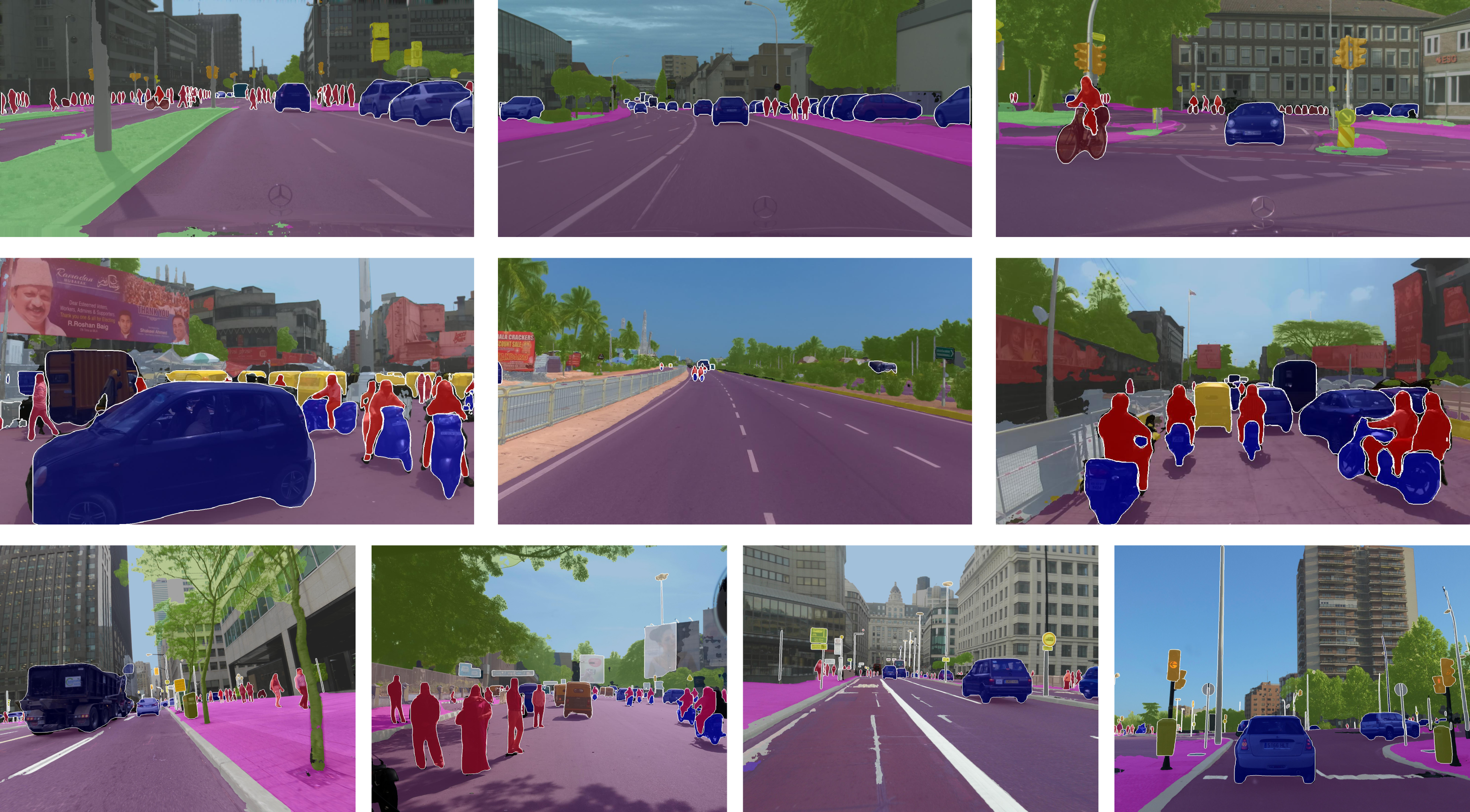}
	\caption{Qualitative results obtained by our proposed combined architecture. Top row: Cityscapes. Middle row: IDD. Bottom row: Vistas. Best viewed in color and with digital zoom.} \label{fig:expresults}
	\vspace{-10pt}
\end{figure*}

We assess the benefits of our proposed network architecture on multiple street-level image datasets, namely Cityscapes~\cite{Cordts2016}, Mapillary Vistas~\cite{Neuhold2017} and the Indian Driving Dataset (IDD)~\cite{Varma19}. 
All experiments were designed to provide a fair comparison between baseline reference models and our proposed architecture design choices. 
To increase transparency of our proposed design contributions, we deliberately leave out model extensions like path aggregation network extensions~\cite{Liu2018,Chen2018}, deformable convolutions~\cite{Dai2017} or Cascade R-CNN~\cite{Cai2018}. We do not apply test time data augmentation (multi-scale testing or horizontal flipping) or explicit use of model ensembles, \etc, as we assume that such bells and whistles approximately equally increase recognition performances for all methods. All models were only pre-trained on ImageNet~\cite{Rus+15}. We use the following terminology in the remainder of this section: \indep~refers to fused, but individually trained models (Fig.~\ref{fig:basic} left) each following the proposed design, and \joint~refers to the unified architecture in Fig.~\ref{fig:basic} (right). 


\paragraph{Model and Training Hyperparameters.}
Unless otherwise noted, we take all the hyperparameters of the instance segmentation branch from~\cite{He2017}. 
These hyperparameters are shared by all the models we evaluate in our experiments.
We initialize our backbone model with weights extracted from PyTorch's ImageNet-pretrained ResNet-50 despite using a different activation function, motivated by findings in our prior work~\cite{RotPorKon18a}. 
We train all our networks with SGD, using a fixed schedule of 48k iterations and learning rate $10^{-2}$, decreasing the learning rate by a factor 10 after 36k and 44k iterations.
At the beginning of training we perform a warm-up phase where the learning rate is linearly increased from $\frac{1}{3}\cdot 10^{-2}$ to $10^{-2}$ in 200 iterations.\footnote{Note that the warm-up phase is not strictly needed for convergence. Instead, we adopt it for compatibility with~\cite{He2017}.}
During training the networks receive full images as input, randomly flipped in the horizontal direction, and scaled such that their shortest side measures $\lfloor 1024\cdot t\rfloor$ pixels, where $t$ is randomly sampled from $[0.5, 2.0]$. Training is performed on batches of 8 images using a computing node equipped with 8 Nvidia V100 GPUs. At test time, images are scaled such that their shortest size measures 1024 pixels (preserving aspect ratio).

\paragraph{Differences with respect to~\cite{Por+19_cvpr}.}
Some scores reported in this document differ from our corresponding CVPR 2019 paper, due to mistakenly copying numbers from the experiment logs and improper handling of \textit{void} labels in Vistas. For fair comparisons please cite the numbers given here.


\subsection{Cityscapes}
\label{ssec:Cityscpes}
Cityscapes~\cite{Cordts2016} is a street-level driving dataset with images from 50 central-European cities. All images were recorded with a single camera type, image resolution of $1024\times 2048$, and during comparable weather and lighting conditions. 
It has a total of 5,000 pixel-specifically annotated images (2,975/500/1,525 for training, validation and test, respectively), and additionally provides 19,998 images forming the \textit{coarse extra} set, where only coarse annotations per image are available (which we have not used in our experiments). Images are annotated into 19 object classes (11 stuff and 8 instance-specific). 

For \indep, we trained each recognition model independently, using the hyperparameter settings described above (again, each with a ResNet-50+FPN backbone), and fused the outputs into a single panoptic prediction as described in Section~\ref{ssec:fusion}. To assess the quality of our obtained panoptic results in terms of standard segmentation metrics, we drop the information about instance identity and retain only the pixel-wise class assignment. By doing so, we can evaluate the quality of \indep in terms of mIoU (mean Intersection-over-Union~\cite{Everingham2010}) against other state-of-the-art semantic segmentation methods. We obtain a result of $75.4$\%, which is comparable or slightly better than $75.2$\% reported in~\cite{Kreso_2017_ICCV} (using a DenseNet-169 backbone), $73.6$\% using DeepLab2 in combination with a ResNet-101 backbone as reported in~\cite{RotNeuKon17cvpr}, or $74.6$\% with a ResNet-152 in~\cite{Wu2016}. Similarly, we can evaluate the instance-segmentation results in terms of AP$_M$ (mean average precision on masks)  by setting to void the stuff pixels. The result of our baseline is $31.9$\%, which is slightly above the reported baseline score in Mask R-CNN~\cite{He2017} (31.5\% w/o COCO~\cite{LinMSCOCO2014} pre-training).

As for the PQ metrics, \indep delivers $\text{PQ}=59.8$\%, PQ$_{\text{St}}=64.5$\%, PQ$_{\text{Th}}=53.4$\% and PQ$^{\dagger}=59.0$\%. We also provide results of \joint~in Tab.~\ref{tab:results}, performing slightly better on both PQ and PQ$^{\dagger}$. 
This is remarkable, given the significantly reduced number of model parameters (see Section~\ref{ssec:computational}) and when assuming that the fusion of individually trained models could lead to an ensemble effect (often deliberately used to improve test results, at the cost of increased computational complexity).

In addition, we show results of jointly trained networks from independent, concurrently appearing works~\cite{Geus18,Li2018,Xiong_UBER_2019,Kirillov19,Yang_Google_2019}, with focus on comparability of network architectures and data used for pre-training. In Tab.~\ref{tab:results} we abbreviate the network backbones as R50, R101 or X71 for ResNet50, ResNet101 or Xception Net71, respectively, and provide datasets used for pre-training (\textbf{I} = ImageNet and \textbf{C} = COCO). All our proposed variants outperform the direct competitors by a considerable margin, \ie our baseline models as well as jointly trained architectures are better.
Finally, the top row in Fig.~\ref{fig:expresults} shows some qualitative seamless segmentation results obtained with our architecture.

\begin{table*}[t]
    \centering
    \resizebox{\textwidth}{!}{
    \begin{tabular}{lll|cccccc|cccccc}
		\toprule
		& & & \multicolumn{6}{c|}{\textbf{Cityscapes}} & \multicolumn{6}{c}{\textbf{Mapillary Vistas}} \\
		Method & Body & Data & PQ & PQ$_{\text{St}}$ & PQ$_{\text{Th}}$ & \PQOurs & AP$_M$ & IoU & PQ & PQ$_{\text{St}}$ & PQ$_{\text{Th}}$ & \PQOurs & AP$_M$ & IoU\\
		\midrule
		de Geus \etal~\cite{Geus18} & R50 & \textbf{I} & - & - & - & - & - & - & 17.6 & 27.5 & 10.0 & - & - & 34.7  \\
		Supervised in~\cite{Li2018Weaklypanoptic} & R101 & \textbf{I} & 47.3 & 52.9 & 39.6 & - & 24.3 & 71.6  & - & - & - & - & - & - \\
		FPN-Panoptic~\cite{Kirillov19} & R50 & \textbf{I} & 57.7 & 62.2 & 51.6 & - & 32.0 & 75.0 & - & - & - & - & - & - \\
		TASCNet~\cite{Li2018} & R50 & \textbf{I+C} & 59.2 & 61.5 & 56.0 & - & 37.6 & 77.8 & 32.6 & 34.4 & 31.1 & - & 18.5 & - \\
		UPSNet~\cite{Xiong_UBER_2019} & R50 & \textbf{I} & 59.3 & 62.7 & 54.6 & - &  33.3 & 75.2  & - & - & - & - & - & - \\
		DeeperLab~\cite{Yang_Google_2019} & X71 & \textbf{I} & 56.3 & - & - & - & - & - & 32.0 & - & - & - & - & 55.3 \\
		\midrule 
		\indep & R50 & \textbf{I} & 59.8 & 64.5 & 53.4 & 59.0 & 31.9 & 75.4 & 37.2 & 42.5 & 33.2 & 38.6 & 16.3 & 50.2 \\
		\rowcolor{mapillarygreen}
		\joint & R50 & \textbf{I} & 60.3 & 63.3 & 56.1 & 59.6 & 33.6 & 77.5 & 37.7 & 42.9 & 33.8 & 39.0 & 16.4 & 50.4 \\
		\bottomrule
	\end{tabular} }
	\caption{Comparison of validation set results on Cityscapes and Vistas with related works. Used network bodies include R101, R50 and X71 for ResNet-101, ResNet-50 and Xception-71, respectively. \textit{Data} indicates datasets used for pre-training where \textbf{I} = ImageNet and \textbf{C} = COCO. All results in [\%]. }
	\label{tab:results}
	\vspace{-10pt}
\end{table*}

\subsection{Indian Driving Dataset (IDD)}
\label{ssec:IDD}
IDD~\cite{Varma19} was introduced for testing perception algorithm performance in India. It comprises 10,003 images from 182 driving sequences, divided in 6,993/981/2,029 images for training, validation and test, respectively. Images are either of 720p or 1080p resolution and were obtained from a front-facing camera mounted on a car roof. The dataset is annotated into 26 classes (17 stuff and 9 instance-specific), and we report results for \textit{level 3} labels. 

Following the same procedure described for Cityscapes, we evaluate \indep~
on segmentation and instance segmentation yielding $\text{IoU}=68.0$\% and AP$_M=32.1$\%, respectively. To contextualize the obtained results, the numbers reported as baselines in~\cite{Varma19} for semantic segmentation are $55.4$\% using ERFNet~\cite{Romera18} and $66.6$\% for dilated residual nets~\cite{Yu_2017_CVPR} and again Mask R-CNN for instance-specific segmentation on a ResNet-101 body yielding AP$_M=26.8$\%. Those numbers supposedly belong to the test set, while no numbers are reported for validation. In terms of panoptic metrics, \indep~yields $\text{PQ}=47.2$\%, PQ$_{\text{St}}=46.6$\%, PQ$_{\text{Th}}=48.3$\% and PQ$^{\dagger}=48.8$\%. For \joint~we obtain $\text{PQ}=46.9$\%, PQ$_{\text{St}}=45.9$\%, PQ$_{\text{Th}}=48.7$\%, PQ$^{\dagger}=48.6$\%, AP$_M=29.8$\% and $\text{IOU}=68.2$\%. In the key metrics PQ and PQ$^{\dagger}$ the results differ by $\leq0.3$ points, and we again stress that the numbers for \joint~are provided from a network with significantly less parameters.


The middle row in Fig.~\ref{fig:expresults} shows seamless segmentation results obtained by our combined architecture.

\subsection{Mapillary Vistas}\label{ssec:vistas}
Mapillary Vistas~\cite{Neuhold2017} is one of the richest, publicly available street-level image datasets today. It comprises 25k high-resolution (on average 8.6 MPixels) images, split into sets of 18k/2k/5k images for training, validation and test, respectively. We only used the training set during model training while evaluating on the validation set. Vistas shows street-level images from all over the world, with images captured from driving cars as well as pedestrians taken them on a sidewalk. It also has large variability in terms of weather, lighting, capture time during day and season, sensor type, \etc., making it a very challenging road scene segmentation benchmark.
Accounting for this, we modify some of the model hyper-parameters and training schedule as follows: we use anchors with aspect ratios in $\{0.2, 0.5, 1, 2, 5\}$ and area $(2\times D)^2$, where $D$ is the FPN level downsampling factor; we train on images with shortest side scaled to $\lfloor 1920\,t\rfloor$, where t is randomly sampled from $[0.8, 1.25]$; we train for a total of 192k iterations, decreasing the learning rate after 144k and 176k iterations.

The results obtained with \indep~and \joint~ are given in Tab.~\ref{tab:results}. The joint architecture besides using significantly less parameters achieves also slightly better results compared to the independently trained models (+$0.5$\% PQ and +$0.4$\% PQ$^\dagger$). Compared to other state-of-the-art methods, we obtain $+5.7$\% and $+5.1$\% PQ score over DeeperLab~\cite{Yang_Google_2019} and TASCNet~\cite{Li2018}, respectively, despite using a weaker backbone compared to DeeperLab and not pre-training on COCO as opposed to TASCNet. 

Finally, we show seamless scene segmentation results in the bottom row of Fig.~\ref{fig:expresults}.

\subsection{Computational Aspects}
\label{ssec:computational}
We discuss computational aspects when comparing the two individually trained recognition models against our combined model architecture. 
When fused, the two task-specific models have $\approx78.06M$ parameters, \ie \mbox{$\approx51.8\%$} more than our combined architecture (\mbox{$\approx51.43M$}). The majority of saved parameters belong to the backbone. The amount of computation is similarly reduced, \ie the combined, independently trained models require $\approx 50.4\%$ more FLOPs due to two inference steps per test image. In absolute terms, the individual models require $\approx 0.864$ TFLOP while our combined architectures requires $\approx 0.514$ TFLOP on $1024\times 2048$ image resolution, respectively.

\subsection{Ablation of the New Pooling in DeeplabV3}
In Section~\ref{sec:sem}, we propose a modification of the DeepLabV3 global pooling operation in order to limit its extent and preserve translation equivariance. Specifically, DeepLabV3 applies global average spatial pooling to the backbone outputs and replicates the outcome to each spatial location. Instead, we replace the global pooling operation with average pooling with a fixed kernel size (in our experiments we use $96\times 96$ corresponding to an image area of $768\times 768$), with stride $1$ and no padding. This is equivalent to a convolution with a smoothing kernel that yields a reduced spatial output resolution. To recover the original spatial resolution we apply padding with replicated boundary. Our motivation is twofold: i) if we train without limiting the input size (\ie~without crops) the context that is captured by global pooling is too wide and degrades the performance compared to having a reduced context as the one captured by our modification and ii) in case one trains with crops, the use of global pooling also at test time breaks translation equivariance and induces a substantial change of feature representation between test and training time, while our modification is not affected by this issue.
To give an experimental evidence of the former motivation, we trained a segmentation model with the DeepLabV3 head and ResNet-50 backbone on Cityscapes at full resolution by keeping the original DeepLabV3 pooling strategy yielding $72.7\%$ mIoU, whereas the same network and training schedule yields $74.5\%$ (+1.8\% absolute) with our new pooling strategy. To support the second motivation, we trained our baseline segmentation model with random $768\times 768$ crops using the standard DeepLabV3 pooling strategy, yielding $73.3\%$ mIoU, whereas the same network with our pooling strategy yields $74.2\%$ (+0.9\% absolute).

\section{Conclusions}
In this work we have introduced a novel CNN architecture for producing \textit{seamless scene segmentation results}, \ie semantic segmentation and instance segmentation modules jointly operating on top of a single network backbone. We depart from the prevailing approach of training individual recognition models, and instead introduce a multi-task architecture that benefits from interleaving network components as well as a novel segmentation module. We also revisit the panoptic metric used to assess combined segmentation and detection results and propose a relaxed alternative for handling stuff segments. Our findings include that we can generate state-of-the-art recognition results that are significantly more efficient in terms of computational effort and model sizes, compared to combined, individual models.

{\small

}

\end{document}